\documentclass{article}

\usepackage{arxiv}
\usepackage[utf8]{inputenc} 
\usepackage[T1]{fontenc}    
\usepackage{hyperref}       
\usepackage{url}            
\usepackage{booktabs}       
\usepackage{amsfonts}       
\usepackage{nicefrac}       
\usepackage{microtype}      
\usepackage{lipsum}		
\usepackage[pdftex]{graphicx}
\usepackage{subcaption}
\usepackage{doi}
\usepackage[load-configurations=version-1]{siunitx} 
\usepackage{placeins}
\usepackage{chngcntr}
\usepackage{blindtext}
\usepackage{enumitem}
\usepackage{mdwlist}
\usepackage{tabularx}
\usepackage[font=footnotesize]{caption}

\title{Early prediction of respiratory failure \linebreak
      in the intensive care unit}

\author{Matthias H\"user \thanks{These
authors contributed equally} \\
	Department of Computer Science \\
	ETH Z\"urich\\
	Zürich, Switzerland \\
	\And
    Martin Faltys \footnotemark[1] \\
    Department of Intensive Care Medicine \\
    University Hospital \\
	University of Bern \\
	Bern, Switzerland \\
	\And
    Xinrui Lyu \footnotemark[1] \\
	Department of Computer Science\\
	ETH Z\"urich\\
	Zürich, Switzerland \\
	\And
    Chris Barber\\
	Department of Computer Science\\
	ETH Z\"urich\\
	Zürich, Switzerland \\
	\And
    Stephanie L. Hyland \\
	Microsoft Research\\
	Cambridge, United Kingdom\\
	\And
    Tobias M. Merz \thanks{Correspondence to 
   \texttt{TobiasM@adhb.govt.nz} and \texttt{raetsch@inf.ethz.ch}} \\
	Cardiovascular Intensive Care Unit\\
	Auckland City Hospital\\
	Auckland, New Zealand\\
	\And
    Gunnar R\"atsch \footnotemark[2] \\
	Department of Computer Science\\
	ETH Z\"urich\\
	Zürich, Switzerland \\
}

\date{}

\renewcommand{\headeright}{}
\renewcommand{\undertitle}{}
\renewcommand{\shorttitle}{Early prediction of respiratory failure in the intensive care unit}

\hypersetup{
pdftitle={Early prediction of respiratory failure in the intensive care unit},
pdfsubject={},
pdfauthor={},
pdfkeywords={},
}

\begin{document}
\maketitle

\begin{abstract}
\small The development of respiratory failure is common among patients in intensive care units (ICU). Large data quantities from ICU patient monitoring systems make timely and comprehensive analysis by clinicians difficult but are ideal for automatic processing by machine learning algorithms. Early prediction of respiratory system failure could alert clinicians to patients at risk of respiratory failure and allow for early patient reassessment and treatment adjustment. 
We propose an early warning system that predicts moderate/severe respiratory failure up to 8 hours in advance. Our system was trained on HiRID-II, a data-set containing more than 60,000 admissions to a tertiary care ICU. An alarm is typically triggered several hours before the beginning of respiratory failure. Our system outperforms a clinical baseline mimicking traditional clinical decision-making based on pulse-oximetric oxygen saturation and the fraction of inspired oxygen. To provide model introspection and diagnostics, we developed an easy-to-use web browser-based system to explore model input data and predictions visually. 
\end{abstract}


\section{Introduction}

Respiratory failure is a source of considerable mortality
and morbidity in ICU patients. Recognizing 
patients that are at risk of acute respiratory failure 
early could help clinicians to expedite the commencement of appropriate treatment. 
Lung function is clinically evaluated using the ratio of blood oxygen partial pressure (Pa$\text{O}_2$) and fraction of inspired oxygen (Fi$\text{O}_2$), commonly referred to as P/F ratio~\cite{bernard1994american}. 
A healthy person breathing room air is expected to have a P/F ratio of ~475 mmHg (Pa$\text{O}_2$: $\sim$100 mmHg, room air Fi$\text{O}_2$: 21 \%). 
Current medical literature defines respiratory failure in three stages \cite{ARDS_Definition_Task_Force2012-hl}:
\begin{itemize}
    \item \textbf{Mild}: 200 mmHg $\leq$ P/F $<$ 300 mmHg
    \item \textbf{Moderate}:100 mmHg $\leq$ P/F $<$ 200 mmHg
    \item \textbf{Severe}: P/F $<$ 100 mmHg 
\end{itemize}

As of yet few approaches using machine learning models to predict respiratory failure in the ICU have been reported. 
Ding et al. used a random forest classifier to predict patients' risk of developing acute respiratory distress syndrome from information derived from 42 variables collected during the first day of ICU admission \cite{ding2019predictive}, and hence do not provide real-time predictions. 
Kim et al. proposed FAST-PACE which is based on an LSTM to predict cardiac arrest and respiratory failure 1-6 h prior to adverse clinical events \cite{kim2019predicting}, and use a small set of only 9 features, potentially limiting model performance. 
In this work, we propose an \textbf{E}arly \textbf{W}arning \textbf{S}ystem (EWS) that can be used to continuously monitor the patients' respiratory state and alert the clinicians when respiratory failure is impending, using a comprehensive feature set from 25 relevant clinical variables. 
\section{The HiRID-II dataset}
\label{sec:hirid2}

A \textbf{Hi}gh time \textbf{R}esolution \textbf{I}CU \textbf{D}ataset (HiRID) was curated to train an EWS for circulatory system failure in \cite{hyland2020early}. The temporal resolution of vital sign data in HiRID is higher than that in MIMIC-III \cite{johnson2016mimic} and eICU \cite{pollard2018eicu}. 
The high-resolution data facilitates more frequent predictions of the patient state, which can help ICU clinicians to make critical decisions more rapidly.
In this work, we used HiRID-II, an extension of HiRID by including more patient data from 2016 to 2019, to train a machine learning model to perform early prediction of respiratory system deterioration.
After the filtering steps shown in Supplementary Figure \ref{fig:patient_filtering} were performed, 62551 ICU patient admissions remained.
The statistics of development data from HiRID-II are shown in Supplementary Figure \ref{fig:training_statistics}.
We summarized 322 meta-variables using 899 recorded variables in the database, 189 more than those used in \cite{hyland2020early}. 
Each meta-variable was derived from recorded variables with similar medical concepts.
\begin{figure}[!ht]
    \centering
    \includegraphics[width=0.9\columnwidth]{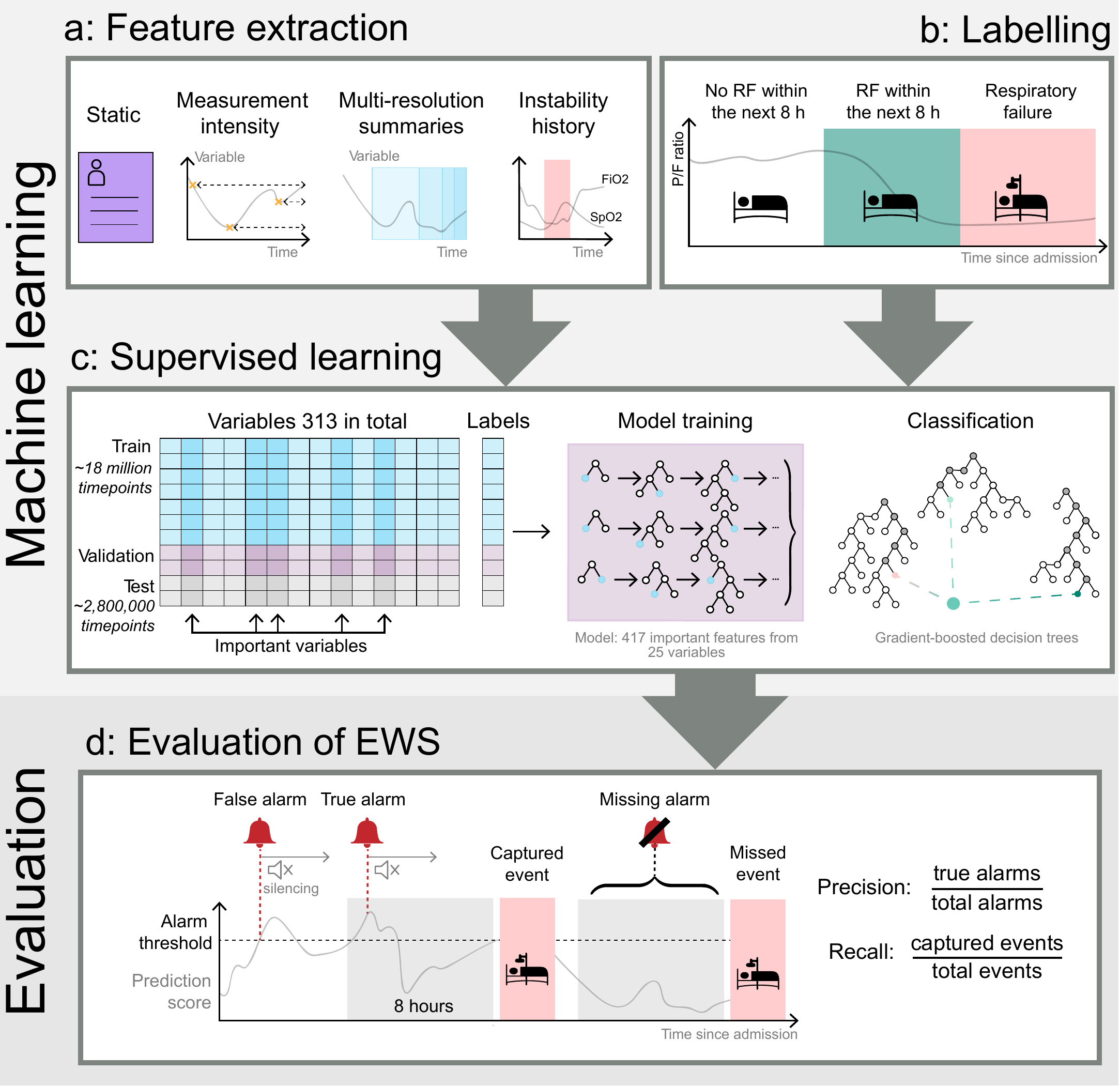}
    \caption{Workflow diagram of the respiratory EWS.}
    \label{fig:framework}
\end{figure}

\section{Continuous estimation of P/F ratio }

The definition of respiratory failure depends on the availability of a current Pa$\text{O}_2$ and Fi$\text{O}_2$ value. 
To measure Pa$\text{O}_2$, an arterial blood sample (ABGA) of the patient has to be drawn and processed. 
Therefore, Pa$\text{O}_2$ measurements are only available at intervals determined by the measurement frequency. 
For a continuous assessment of patient respiratory state using P/F ratio, estimates of Pa$\text{O}_2$ values have to be used when its measurements are not available. 
In the clinical setting, continuously monitored pulse oximetry derived haemoglobin oxygen saturation (Sp$\text{O}_2$) can be used to estimate the current Pa$\text{O}_2$ value \cite{Brown2017-je, Ellis1989-sq, Severinghaus1979-tt}. 
A comparison of different existing models shows that the non-linear parametric model by Ellis \cite{Ellis1989-sq, Severinghaus1979-tt} performs best.
We were able to further improve continuous Pa$\text{O}_2$ estimation using neural networks (NNs).
Two Pa$\text{O}_2$ estimation models were developed: 1) a basic NN relying only on Sp$\text{O}_2$ measurements as input (Sp$\text{O}_2$-NN); 2) a more complex NN that also takes the last ABGA measurement into account (Full-NN). 
\begin{table}[!ht]
    \caption{Area under the receiver operating characteristic (AUROC) and median absolute Pa$\text{O}_2$ estimation error of the two NNs and the parametric non-linear baseline (pnl-baseline) \cite{Ellis1989-sq}. The AUROC shows the performance of the models in detecting P/F ratios $\leq$ 200 mmHg. The best results per category are presented in \textbf{bold}, the second best in \textit{italic}. \vspace{2mm}}
    \label{tab:pao2_pred_condensed}
    \centering
    {\small
        \begin{tabular}{lcccc}
            \toprule
            Range Sp$\text{O}_2$ (\%) & n & pnl-baseline & Sp$\text{O}_2$-NN & Full-NN \\
            \midrule
            Overall & 20994 & {10.1} & \textit{9.6} & \textbf{9.2} \\
            90-96 & 7945 & {6.9} & \textit{6.8} & \textbf{6.3} \\
            80-90 & 780 & {6.1} & \textit{5.1} & \textbf{4.4} \\
            70-80 & 78 & {19.7} & \textit{10.2} & \textbf{8.2} \\
            \midrule
            AUROC & 12823 & 0.914 & \textit{0.917} & \textbf{0.919} \\
            \bottomrule
        \end{tabular}
    }
\end{table}

Table \ref{tab:pao2_pred_condensed} shows that both NNs outperform the best baseline model in absolute estimation error and discriminating patients with and without moderate/severe respiratory failure. 
We provide more details about the proposed NNs in Appendix \ref{sec:pao2_model}.

For the calculation of the P/F ratio, estimates of Fi$\text{O}_2$ values are also necessary. Three situations need to be distinguished: 1) the patient is breathing ambient air, i.e. $\text{FiO}_2=21 \%$ (the ambient air oxygen fraction); 2) for patients receiving supplemental oxygen Fi$\text{O}_2$ is estimated using the lookup table by \cite{Wettstein2005-ol} (see Supplementary Table \ref{tab:supp_fio2_mapping}); 3) for patients on mechanical ventilation Fi$\text{O}_2$ is controlled by the ventilator and its value is recorded in the data.

Here we are interested in predicting respiratory failure of only the moderate/severe level as previously defined.
Therefore, a time point $t$ is labeled as positive for respiratory failure if the following conditions are true for two thirds of a two-hour forward window (see Figure \ref{fig:framework}b):
\begin{itemize}
    \item P/F ratio $<$ 200 mmHg; 
    \item for mechanically ventilated patients: the positive end expiratory pressure (PEEP) is $\geq$ 5 mmHg.
\end{itemize}

To remove spurious detections, we merged successive events with a gap of at most 1h into one event, and deleted short events of length at most 2h.
Our model then predicts the patient being in respiratory failure in the next 8h according to the above definition, given they are currently stable, which implements an early warning system.

\section{Early warning system for low P/F ratio}

As an initial step, the 20 most influential variables for predictions were retrieved on a basic feature
set including measurement intensity and the current parameter value using SHAP value analysis \cite{lundberg2017consistent}, and 
augmented with clinically important variables, yielding $n_\text{var}=25$ variables
used for model development.

To give our model a comprehensive view of the patient state the following feature classes were extracted (see Figure \ref{fig:framework}a) on these variables:
\begin{itemize}
    \item {\bf Current value:}~The current parameter value.
    \item {\bf Multi-resolution summary:}~The summary functions $\{\text{mean}, \text{std}, \text{trend}, \text{min}, \text{max}\}$ were computed over the last 8 hours as well as the entire stay up to now.
    \item {\bf Measurement intensity:}~The time to last real measurement, measurement density in the last 8 hours as well as during the entire stay up to now.
    \item {\bf Instability history:}~If applicable for a variable, up to 3 severity levels were annotated using prior medical knowledge. The fraction of time spent in each severity state over the last 8 hours as well as over the entire stay up to now was extracted.
\end{itemize} 
Static variables were used directly as a feature. 

These features were then passed to an ensemble
of decision trees implemented in \texttt{LightGBM} \cite{ke2017lightgbm}, shown in Figure \ref{fig:framework}c. Trees were added
to the ensemble until performance did not improve for 50 epochs in the
validation set, early stopping the training process. Prediction scores were generated for patients
in the test set, and form the basis of the early warning system.

\begin{figure}[!ht]
    \centering
    \begin{subfigure}[b]{0.4\textwidth}
    \includegraphics[width=\textwidth]{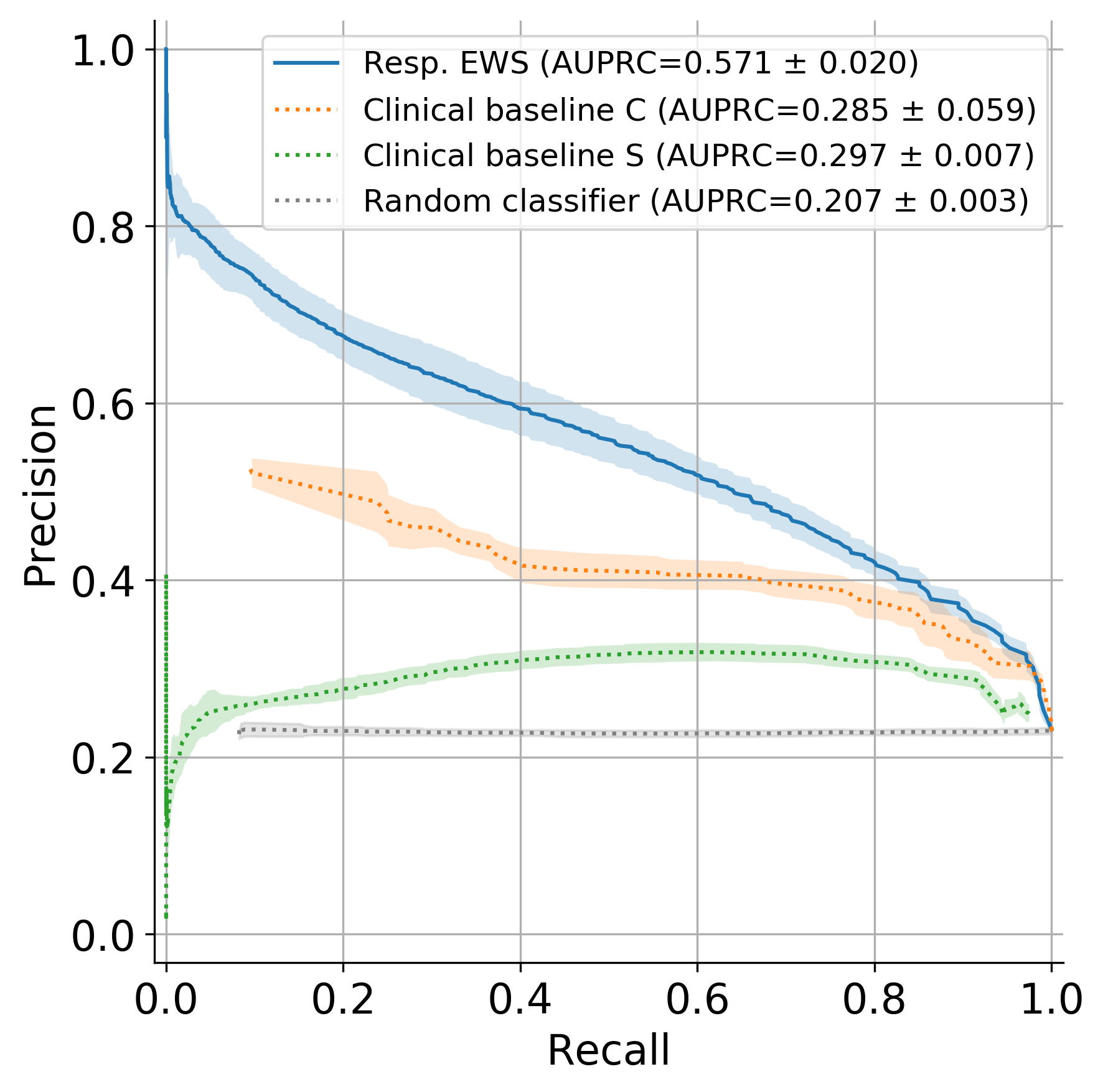}
    \caption{~}
    \label{fig:main-event-based-result}
    \end{subfigure}
    ~
    \begin{subfigure}[b]{0.58\textwidth}
    \includegraphics[width=\textwidth]{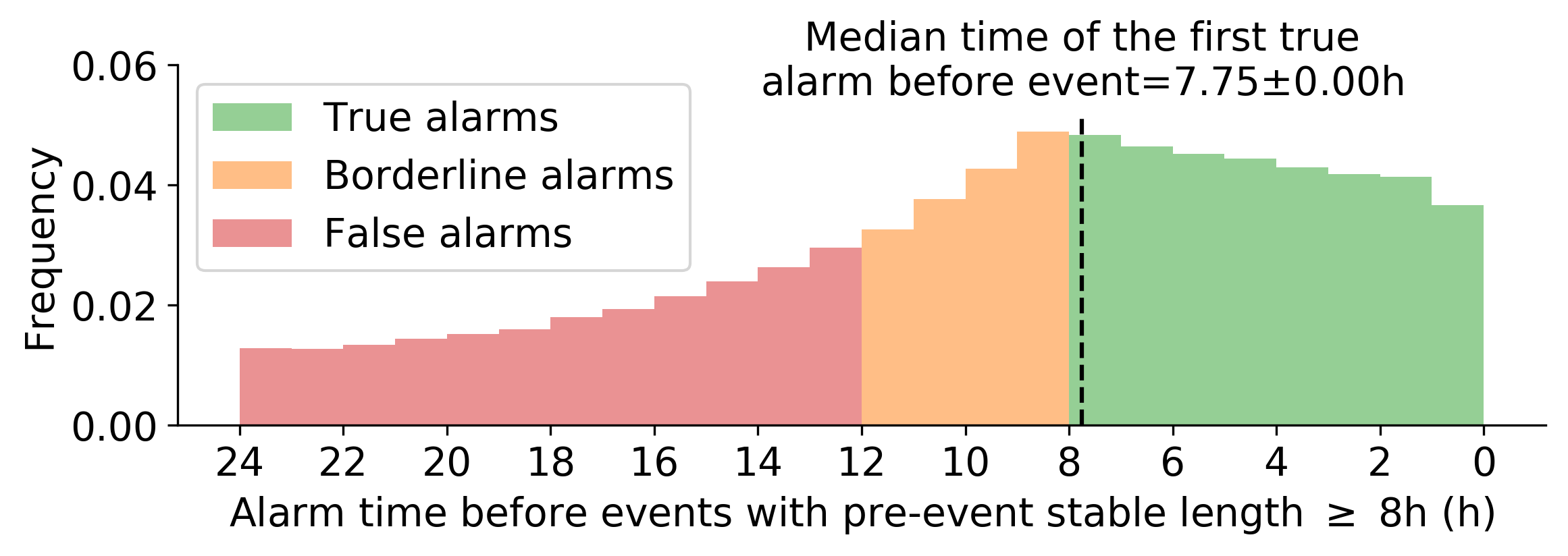}
    \includegraphics[width=\textwidth]{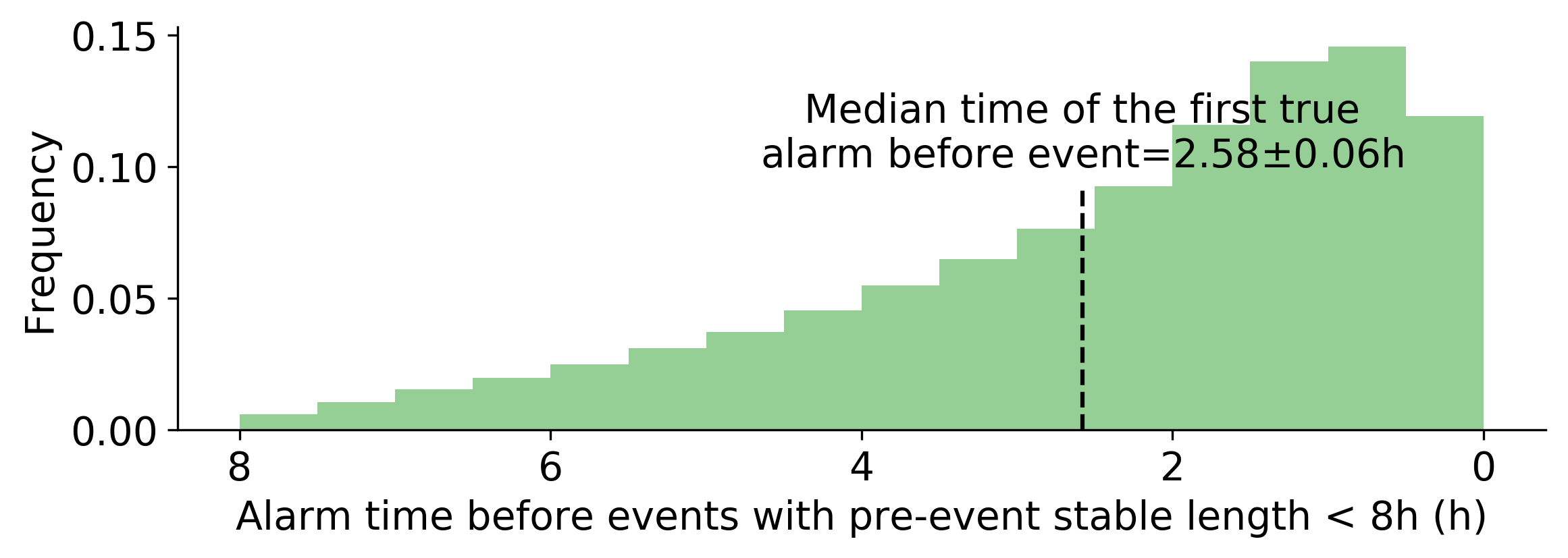}
    \caption{~}
    \label{fig:first_alarm_hist}
    \end{subfigure}
    \caption{(a) Event-based precision-recall curve of EWS for acute respiratory failure, two clinical baselines and a random classifier. \textbf{Clinical baseline S} is a simple baseline solely based on Sp$\text{O}_2$ lower than the specified threshold. \textbf{Clinical baseline C} is a more complex baseline that uses a single decision tree trained on two basic features, the current estimate of Sp$\text{O}_2$ and Fi$\text{O}_2$, with at most 32 leaves. \textbf{Random classifier} shows the event prevalence. Both baselines and the random classifier also use the same alarm silencing strategy as our method. Error bands show the standard deviation of model performance in 5 separate data splits, similar to \cite{hyland2020early}. (b) Distribution of alarm time before respiratory failure. The median time of the first true alarm before all events irrespective of the pre-event stability length is 3.75$\pm$0.24 h.} 
\end{figure}

The conventional way of evaluating the performance of machine learning models
for time-series is using a time-point based precision-recall curve. 
This is less suitable for clinical settings because it is equivalent to evaluating a system that generates frequent alarms, which can cause alarm fatigue that should be avoided in healthcare settings \cite{erci2011top10harzards}.
We used the alarm silencing strategy proposed by Hyland et al.\cite{hyland2020early}, which silences further potential alarms within a specified period of time after the model output triggers an alarm, hence reducing alarm fatigue.
In our evaluation, we use 30 minutes as the alarm silencing time.
The same event-based precision and recall definition proposed in \cite{hyland2020early} was adopted (see Figure \ref{fig:framework}d).

Figure~\ref{fig:main-event-based-result} shows that our respiratory EWS outperforms the two clinical baselines significantly, and that it only generates 3 false alarms out of 5 alarms at 80\% recall, with the first alarm alerting the clinicians to future respiratory failure occurring on average approximately 4 hours before the actual event, given the event was detected (see Figure~\ref{fig:first_alarm_hist}).
The average number of alarms within the 8-hour window prior to detected respiratory failure events is 7.4,  meaning that the systems will keep informing the clinicians roughly every hour in this period before respiratory failure.

\section{ICU monitor}
To aid model introspection and diagnostics we have developed a user customizable, web-based tool for interactive data exploration of the HiRID time-series data, generated features and model predictions. The system supports the manual annotation of arbitrary time series tracks. This gives clinicians the opportunity to visually discover the data and perform actions such as annotation of interesting regions in the data, or annotation of wrong predictions, putting the clinician into the model improvement loop. For more details about the tool, we refer to the Appendix \ref{sec:monitor}.
\section{Conclusion}
In this work, we describe the development of an early warning system for respiratory failure in the ICU setting. In a first step, we successfully derived a non-parametric model for Pa$\text{O}_2$ estimation based on continuously available Sp$\text{O}_2$ measurements and previous ABGA data, which outperforms the best baseline method. The Pa$\text{O}_2$ model allows for the continuous estimation of the patient respiratory state without frequent ABGA, and also enables the continuous state definition needed for training an early warning system for respiratory failure.
Our model predicts respiratory failure in real time and can alert clinicians early with a lower false alarm rate to impending respiratory failure than the clinical baselines. We hypothesize that our system can facilitate early reassessment of a deteriorating patient, enabling the rapid treatment and improving patient outcomes. 
Our approach allows for feature inspection on individual prediction and SHAP value analysis of important predictors, offering additional valuable insights to clinicians. Our visualization tool enables the visual exploration of the relevant variables and the system's predictions.
Future work will investigate alternative definitions of moderate/severe respiratory failure, as well as exploring alternative modeling choices, including different machine learning models.

\section*{Acknowledgments}

This  project  was  supported  by  the  Grant No. 205321\_176005 
of the Swiss National Science Foundation (to TMM/GR) and ETH
core funding (to GR).

\bibliographystyle{plain}
\bibliography{references} 

\begin{thebibliography}{10}
\bibitem{erci2011top10harzards}
ERCI Institute.
\newblock Top 10 technology hazards for 2012: the risks that should be at the
  top of your prevention list.
\newblock {\em Health Devices}, 40(11):358--373, 2011.
\end{thebibliography}

\appendix
\onecolumn
\counterwithin{figure}{section}
\counterwithin{table}{section}

\section{Patient and variable flow chart}
\begin{figure}[h!]
    \centering
    \includegraphics[width=0.8\textwidth]{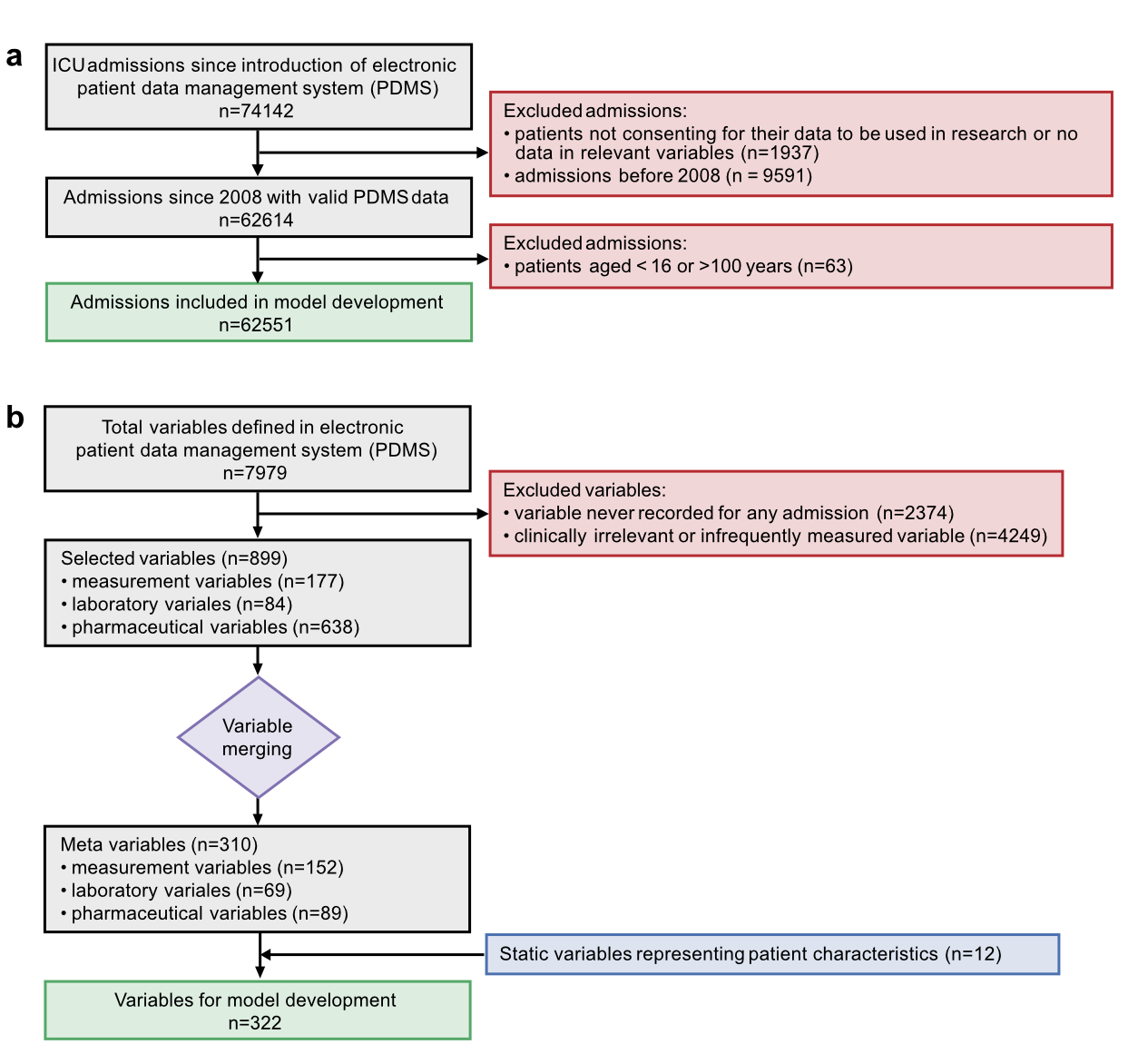}
    \caption{a, Flow chart of the exclusion criteria applied to the HiRID-II patient cohort. b, Flow chart of the exclusion, merging, and post-processing applied to the variables in the patient data management system}
    \label{fig:patient_filtering}
\end{figure}
\clearpage

\section{HiRID-II patient cohort statistics}
\begin{figure}[!ht]
    \centering
    \includegraphics[width=0.7\textwidth]{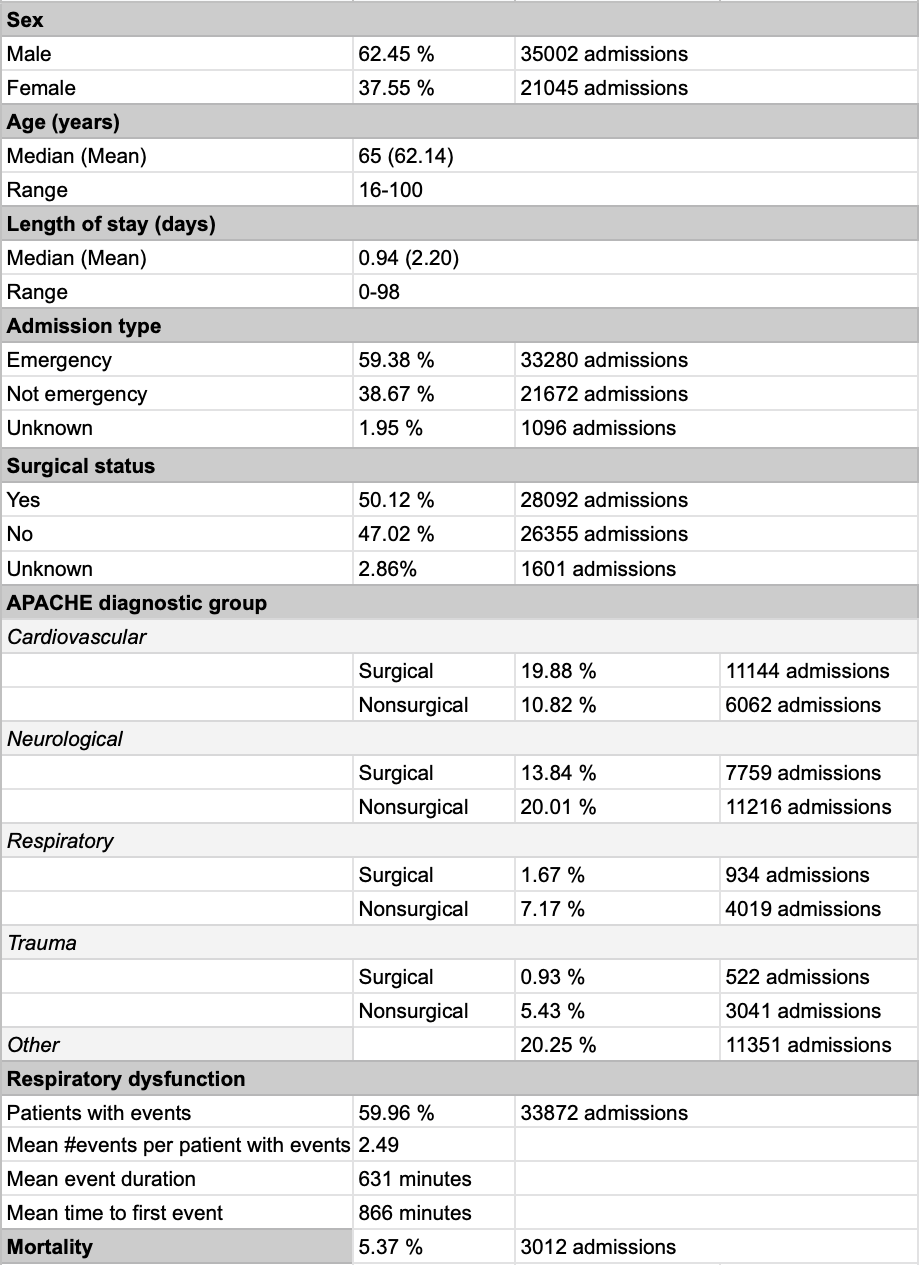}
    \caption{Statistics of the development cohort extracted from the HiRID-II database}
    \label{fig:training_statistics}
\end{figure}
\clearpage

\section{Continuous P/F ratio estimation}
\subsection{Pa$\text{O}_2$ estimation model development}\label{sec:pao2_model}
Both the Sp$\text{O}_2$-NN and Full-NN models are trained using the HiRID dataset \cite{hyland2020early}, which contains 196,148 arterial blood gas analysis (ABGA) samples.
We split the ICU admissions with using 75\%/25\% training/test ratio. 
The training set contains 147'210 ABGA samples from 19'419 ICU admissions, and the test set contains 48'938 ABGA samples from 6'473 ICU admissions.
Before training we excluded a total of 32'210 ABGA samples (30’537 due to value being outside of the normal Pa$\text{O}_2$ range [40, 250] mmHg and 1'673 due to the last ABGA measurement being longer than 24h in the past). 

We applied cross-validation and used grid search in hyperparameter tuning for both proposed NNs.
We summarize the hyperparameters as well as their search space in Table \ref{tab_hyperparm_pao2}, and also show the best hyperparameter set selected for each proposed NN in \ref{tab:opt_hp}. 
The best set of hyperparameters was chosen based on the lowest mean absolute prediction error in the clinically relevant region of Sa$\text{O}_2$ $<$ 96\%. 

For the more complex model, Full-NN, we first train it on a large variable set consisting of Sa$\text{O}_2$, mean etC$\text{O}_2$ (10 min), mean patient body temperature (4 hours) and different measurements from the previous ABGA test (namely Sa$\text{O}_2$, pH, Fi$\text{O}_2$, Pa$\text{O}_2$, Hb,  MetHb, COHb, pC$\text{O}_2$, BE, HC$\text{O}_3$ and lactate).
We then performed a greedy backward selection with 3-fold cross validation to select the most important variables from the initial variable set. 
The four most important variables are the current and the last Sa$\text{O}_2$, as well as Pa$\text{O}_2$ and pH from the last ABGA test. 
It is important to note that the models were trained using Sa$\text{O}_2$ as input for the haemoglobin oxygen saturation, which is the corresponding precise measure of haemoglobin oxygen saturation in an ABGA. As this value hence is also not continuously available, we had to use the less precise, but continuously available Sp$\text{O}_2$ during prediction. The evaluation on the test set was done using Sp$\text{O}_2$ to reflect the use case. All model development was performed with Tensorflow 1.15.
\begin{table}[!ht]
    \caption{Hyperparameters for the proposed NNs and their search space. \textbf{Hidden layer nodes}: each element $n$ in the tuple represents a hidden layer with $n$ nodes. \textbf{Weight correction factor $\gamma$}: let $x_i$ be the Sa$\text{O}_2$ value of the $i$-th training example, and we define $c := |\big\{x_j=x_i|j\in\{1,\ldots,N_\text{train}\}\big\}|$. During training the cost of the $i$-th training example is multiplied by 1/$c^\gamma$.  \textbf{Dropout rate}: the fraction of nodes being randomly dropped out in each hidden layer during training.}
    \label{tab_hyperparm_pao2}
    \begin{center}
        \resizebox{0.8\textwidth}{!}{%
        \begin{tabularx}{\textwidth}{lX}
            \toprule
            \textbf{Hyperparameter} & \textbf{Search space} \\ \midrule 
            Batch size  & \{30, 50, 100, 300, 500\} \vspace{2mm} \\ 
            Hidden layer nodes  & \{(8,8), (16,16), (32,32), (64,64), (128,128), (256,256), (64,128), (128,64), (64,64,64), (64,128,64), (128,128,128), (128,256,128), (256,512,256)\} \vspace{2mm} \\ 
            Weight correction factor ${\gamma}$ & \{None, 0.1, 0.2, 0.33, 0.5, 1\} \vspace{2mm} \\ 
            Learning rate & $[10^{-4}, 10^{-1})$ (10 grid points) \vspace{2mm} \\
            Dropout rate & $[0, 0.5)$ (10 grid points) \\
            \bottomrule
        \end{tabularx}%
        }
    \end{center}
\end{table}
\begin{table}[!ht]
    \caption{The best hyperparameters selected for the two Pa$\text{O}_2$ estimation NNs \vspace{1mm}}
    \label{tab:opt_hp}
    \centering
    \resizebox{0.8\textwidth}{!}{%
        \begin{tabularx}{\textwidth}{lX}
            \toprule
            \textbf{Model} & \textbf{Selected hyperparameters} \\ \midrule 
            Sp$\text{O}_2$-NN & \textbf{\small Batch size:} 50; \textbf{\small Hidden layer nodes:} (64,128,64); \textbf{\small Weight correction factor:} None; \textbf{\small Learning rate:} 0.0001; \textbf{\small Dropout rate:} 0.5 \vspace{2mm} \\
            Full-NN & \textbf{\small Batch size:} 50, \textbf{\small Hidden layer nodes:} (8,8), \textbf{\small Weight correction factor:} 0.2; \textbf{\small Learning rate}: 0.001, \textbf{\small Dropout rate}: None \\
            \bottomrule
        \end{tabularx}%
    }
\end{table}
\newpage
\subsection{Pa$\text{O}_2$ estimation model performance}
\begin{table}[!ht]
    \caption{Area under the receiver operating characteristic (AUROC) and median absolute Pa$\text{O}_2$ estimation error of the two NNs and the parametric non-linear baseline (pnl-baseline) \cite{Ellis1989-sq}. The AUROC shows the performance of the models in detecting P/F ratios $\leq$ 200 mmHg. The best results per category are presented in \textbf{bold}, the second best in \textit{italic}. \vspace{2mm}}
    \label{tab:pao2_pred}
    \centering
    {\small
        \begin{tabular}{lcccc} \toprule
            Range Sp$\text{O}_2$ \% & n & pnl-baseline (iqr) & Sp$\text{O}_2$ NN (iqr) & Full NN (iqr) \\
            \midrule
            0-100 & 20994 & {10.1 (16.1)} & \textit{9.6 (15.0)} & \textbf{9.2 (14.8)} \\
            0-96 & 8813 & {6.8 (9.9)} & \textit{6.7 (10.3)} & \textbf{6.1 (9.8)} \\
            96-100 & 12181 & {13.8 (19.1)} & \textit{12.4 (17.5)} & \textbf{12.3 (17.2)} \\
            90-96 & 7945 & {6.9 (9.9)} & \textit{6.8 (10.2)} & \textbf{6.3 (9.9)} \\
            85-90 & 636 & {5.8 (8.3)} & \textit{5.3 (6.9)} & \textbf{4.4 (6.7)} \\
            80-85 & 144 & {7.2 (15.8)} & \textit{4.8 (11.5)} & \textbf{4.7 (10.4)} \\
            75-80 & 63 & {14.6 (38.2)} & \textit{7.1 (35.9)} & \textbf{6.7 (32.9)} \\
            60-75 & 22 & {29.5 (52.5)} & \textit{17.5 (55.0)} & \textbf{14.0 (52.1)} \\
            \midrule
            auROC & 12823 & 0.914 & \textit{0.917} & \textbf{0.919} \\
           \bottomrule
        \end{tabular}
    }
\end{table}

\subsection{Fi$\text{O}_2$ estimation}
\begin{table}[!ht]
    \caption{ Supplemental Oxygen to Fi$\text{O}_2$ conversion table used for determining the continuous FiO2 estimate.}
    \label{tab:supp_fio2_mapping}
    \begin{center}
        {\small
            \begin{tabular}{ll}
                \toprule
                \textbf{Supp. oxygen [l]} & \textbf{Fi$\text{O}_2$ [\%]} \\
                \midrule
                1  & 26 \\
                2 & 34 \\
                3 & 39 \\
                4 & 45 \\
                5 & 49 \\
                6 & 54 \\
                7 & 57 \\
                8 & 58 \\
                \bottomrule
            \end{tabular}
            \begin{tabular}{ll}
                \toprule
                \textbf{Supp. oxygen [l]} & \textbf{Fi$\text{O}_2$ [\%]} \\
                \midrule
                9 & 63 \\
                10 & 66 \\
                11 & 67 \\
                12 & 69 \\
                13 & 70 \\
                14 & 73 \\
                15 & 75 \\
                $>$15 & 75 \\
                \bottomrule
            \end{tabular}%
        }
    \end{center}
\end{table}

\clearpage

\section{Variable selection and model introspection}

\begin{table}[h!]
\caption{List of the 20 most important variables according to SHAP value analysis, as well as additional variables, which are plausible given prior medical knowledge. The ranking was determined by ranking the features using the mean absolute SHAP value in the validation set, and then ranking the variables by the rank of their most important
feature. The 25 displayed variables formed the basis
of the proposed prediction model. \vspace{2mm}}
\centering
{ \small
\begin{tabular}{ll}
\toprule
\textbf{Rank} & \textbf{Important variable} \\
\midrule
1 & Fi$\text{O}_2$ \\
2 & Sp$\text{O}_2$ \\
3 & Supplemental oxygen \\
4 & Pa$\text{O}_2$ \\
5 & Supplemental Fi$\text{O}_2$ \% \\
6 & Sa$\text{O}_2$ \\
7 & GCS Eye opening \\
8 & GCS Response \\
9 & Peritoneal dialysis  \\
10 & Peak pressure \\
\bottomrule
\end{tabular}
\begin{tabular}{ll}
\toprule
\textbf{Rank} & \textbf{Important variable} \\
\midrule
11 & Spontaneous breathing? \\
12 & Admission origin \\
13 & GCS Motor \\
14 & Weight \\
15 & Ventilator mode group \\
16 & RASS \\
17 & Patient age \\
18 & Extubation time-point \\
19 & Tracheotomy state \\
20 & ST2 (ECG) \\
\bottomrule
\end{tabular}
\begin{tabular}{l}
\toprule
\textbf{Clinically identified variables} \\
\midrule
Respiratory rate \\
PEEP \\
OUT \\
Current Fi$\text{O}_2$ estimate \\
Ventilation state \\
\bottomrule
\end{tabular} }
\label{tab:top_vars}
\end{table}

\begin{table}[h!]
\caption{Most important features identified by SHAP analysis
         in the validation set of the 6 splits. The ranking was obtained
         by first finding the mean of absolute SHAP values for all samples
         in the validation set of a split, and then averaging this mean
         across all splits. \vspace{2mm}
}
\centering
{\small
\begin{tabular}{ll}
\toprule
\textbf{Rank} & \textbf{Important feature} \\
\midrule
1 & Sp$\text{O}_2$ (Mean 8h)  \\
2 & Fi$\text{O}_2$ estimate (Current value)  \\
3 & Fi$\text{O}_2$ estimate (Max 8h) \\
4 & Fi$\text{O}_2$ estimate (Mean 8h)\\
5 & Fi$\text{O}_2$ estimate (Mean entire stay) \\
6 & Sp$\text{O}_2$ (Current value) \\
7 & Sp$\text{O}_2$ (Mean entire stay) \\
8 & Ventilator peak pressure (Current value) \\
9 & Sp$\text{O}_2$ (Instable density of L1 90-94 \% 8h)  \\
10 & Fi$\text{O}_2$ (Instable density of L1 30-40 \% 8h) \\
11 & Peritoneal dialysis (Time to last ms.) \\
12 & Fi$\text{O}_2$ estimate (Std. entire stay) \\
13 & Pa$\text{O}_2$ (Current value)  \\
14 & Supplemental oxygen (Instable density of L1 2-4 l/min 8h)  \\
15 & GCS Eye opening (Current value) \\
16 & Pa$\text{O}_2$ (Min 8h) \\
17 & Supplemental Fi$\text{O}_2$ \% (Instable density of L1 21-40 8h)  \\
18 & Peritoneal dialysis (Meas. density entire stay) \\
19 & Fi$\text{O}_2$ (Current value) \\
20 & Patient age \\
\bottomrule
\end{tabular}
}
\label{tab:top_vars}
\end{table}

\clearpage
\section{ROC-based performance of the prediction score}

For completeness, we also display an evaluation of the 
proposed Resp EWS prediction score using the classical way
of time-point based evaluation using ROC analysis. Sensitivity
is the fraction of correctly retrieved time points at which
an alarm is mandated, in the 8 hours before respiratory failure.
Specificity is the fraction of time-points at which no alarm 
should be produced, in which indeed none is produced. Optimal 
performance under this evaluation implies producing an alarm
every 5 minutes.

\begin{figure}[h!]
    \centering
    \includegraphics[width=0.5\textwidth]{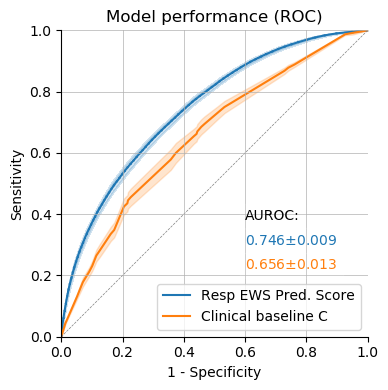}
    \caption{ROC curve of the Resp EWS prediction score underlying
             the proposed alarm system.}
    \label{fig:auroc_results}
\end{figure}
\clearpage

\section{ICU monitor: patient data visualization}\label{sec:monitor}

We have developed a flexible and dynamic web-based viewer for the
HiRID-II time-series data, as well as auxiliary data generated during experiments. Arbitrary channels can be plotted in vertically arranged line or scatter plots, where the x-axes are the date-times of the measurements, and kept synchronized during interactive/animated zoom and pan operations. We use the tool for the exploration and inspection of data across many patients, and for the overlay and evaluation of derived events and quantities relevant to experimentation, like event annotations or machine
learning labels. We plan to share the tool under an open source license, and we describe some of its current features below.

\begin{figure}[ht!]
    \centering
    \includegraphics[width=\textwidth]{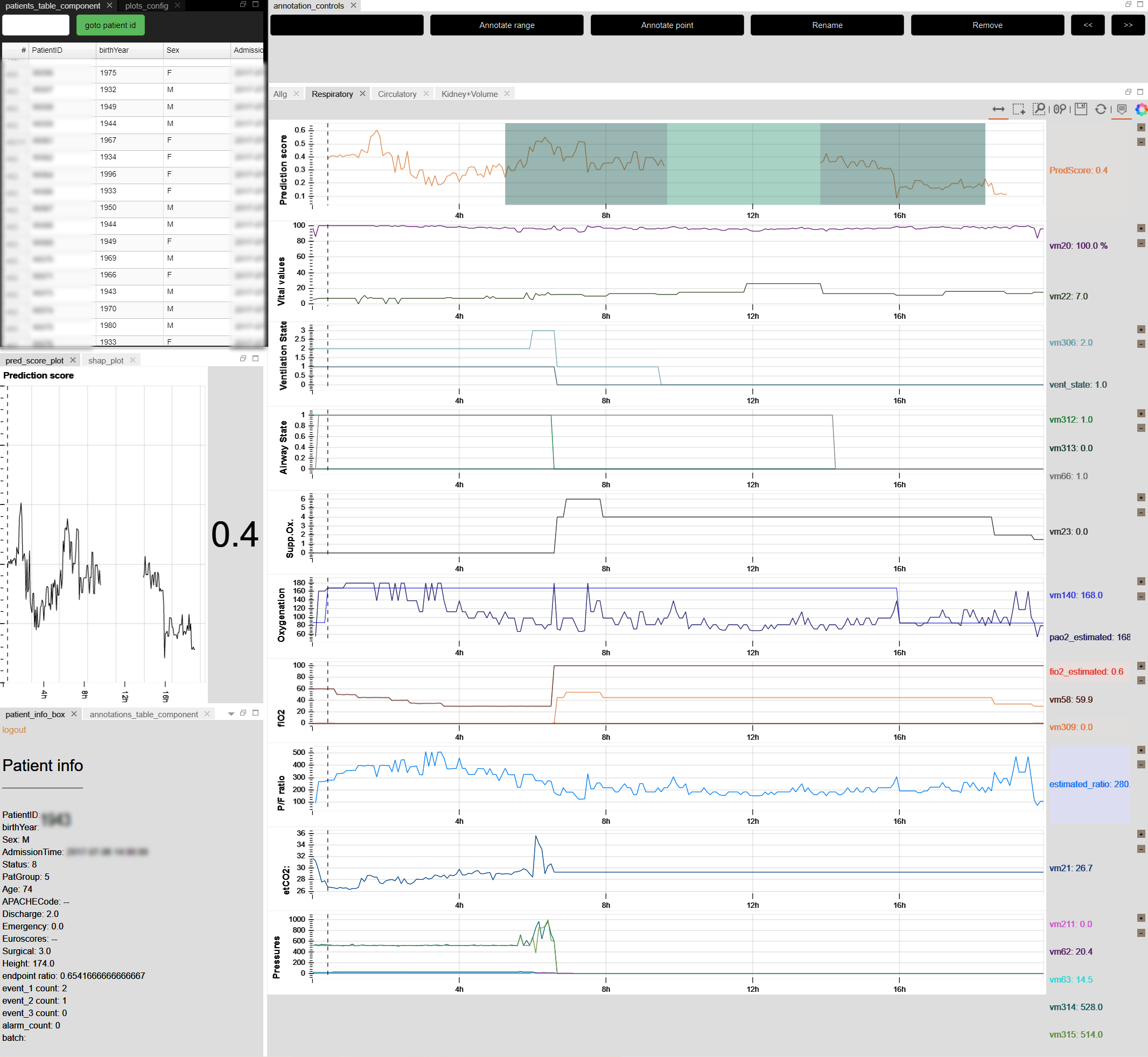}
    \caption{\small Illustrative print screen of the newly developed ICU monitor tool for data inspection. The topmost graph is displaying the models' predicted score. The underlaid regions mark event levels of different severity. The light green region represents a respiratory failure state, therefore no predictions are produced during this period. The other graphs show a customizable selection of the variables. Regions of the ICU monitor displaying patient ID and exact dates are blurred-out.}
    \label{fig:auroc_results}
\end{figure}

\subsection{Visualization features}

\begin{itemize*}
  \item Arbitrary time-resolution (x-axes ticks intelligently adjust to different time scales)
  \item Irregular time-steps (for example, we overlay post-processed data at a regular time interval on top of source data with irregular time steps)
  \item Missing data (missing points are not rendered)
  \item Auto-scaling y-ranges, overridable by configuration
  \item Inspect exact values by hover tooltip, or by moving a ``current timestep'' cursor
  \item Auto color allocation per channel, and readouts with channel label and unit, according to current cursor position
  \item Dynamically sizable heights of individual plots
  \item Multiple channels per plot, up to 2 y-axes per plot
  \item Distribution of plots to separate named tabs
  \item Dynamic resize and popout of tabs (e.g. for viewing on a second monitor)
  \item Distribution of channels to plots, and plots to tabs are both reconfigurable ``live'' without restarting/reloading
  \item Full sortable table of patients, click or type in patient ID to load a patient
  \item Patient info listing, with list of available channels available for use in plot configuration
\end{itemize*}

\subsection{Event annotation features}

\begin{itemize*}
  \item Overlay of ``annotations'' / ``events'' for time ranges or time points
  \item Live modification of annotation/event color per annotation/event type, via the UI
  \item Events loaded statically from data files, or annotations added interactively via UI
  \item Keyboard shortcuts for streamlined use of mouse for both annotation time range selection as well as zoom \& pan
  \item Display of annotation/event labels/names on the plots
  \item JSON-schema defined schema per annotation/event type, allowing entry of structured metadata per annotation/event in the UI
  \item Display and edit metadata in a dynamic form in the UI, and view metadata in special hover tooltip mode
  \item Dump metadata of annotations/events to JSON via predefined route/URL
  \item Sortable and filterable table of annotations/events, selecting a row automatically pans the x-axis such that the annotation/event comes into view
  \item Chronological navigation of annotations/events using the arrow keys
  \item Filtering annotations/events to only certain types of interest (filtered types will be hidden from view and be skipped when navigating chronologically)
  \item Selection of annotations/events by mouse-click, synchronized to the table, and supporting unambiguous behavior for selection on overlapping annotations/events
  \item Keyboard shortcuts for remove/rename annotation
\end{itemize*}

\end{document}